\documentclass[a4paper,journal,twoside,web]{ieeecolor}
\usepackage{tmi}
\usepackage{cite}
\usepackage{amsmath,amssymb,amsfonts}
\usepackage{algorithmic}
\usepackage{graphicx}
\usepackage{textcomp}

\def\BibTeX{{\rm B\kern-.05em{\sc i\kern-.025em b}\kern-.08em
    T\kern-.1667em\lower.7ex\hbox{E}\kern-.125emX}}
\usepackage{caption}
\usepackage{url}

\newcommand{\norm}[1]{\left\lVert#1\right\rVert}
\usepackage{yhmath}

\usepackage{amsthm}
\usepackage{multirow}

\theoremstyle{definition}
\newtheorem*{definition}{Definition}

\begin{document}
\title{Deep Hypothesis Tests Detect Clinically Relevant Subgroup Shifts in Medical Images}
\author{Lisa M. Koch, Christian M. Schürch, Christian F. Baumgartner, Arthur Gretton, Philipp Berens 
%
%
\thanks{Submitted for review on 8 March 2023. This work was supported by the German Science Foundation (BE5601/8-1 and the Excellence Cluster 2064 ``Machine Learning --- New Perspectives for Science'', project number 390727645), the Carl Zeiss Foundation in the project ``Certification and Foundations of Safe Machine Learning Systems in Healthcare'' and the Hertie Foundation. }
\thanks{L. M. Koch (e-mail: lisa.koch@uni-tuebingen.de) and P. Berens are with the Hertie Institute for Artificial Intelligence in Brain Health, the Institute of Ophthalmic Research and the Tübingen AI Center, University of T{\"u}bingen, Germany.}
\thanks{C. M. Sch{\"u}rch is with the Department of Pathology and Neuropathology, University Hospital T{\"u}bingen and Comprehensive Cancer Center T{\"u}bingen, Germany.}
\thanks{C. F. Baumgartner is with the Cluster of Excellence Machine Learning: New Perspectives for Science, University of T{\"u}bingen, T{\"u}bingen, Germany.}
\thanks{A. Gretton is with the Gatsby Computational Neuroscience Unit, University College London, United Kingdom.}}
\maketitle

\begin{abstract}
Distribution shifts remain a fundamental problem for the safe application of machine learning systems. If undetected, they may impact the real-world performance of such systems or will at least render original performance claims invalid. In this paper, we focus on the detection of subgroup shifts, a type of distribution shift that can occur when subgroups have a different prevalence during validation compared to the deployment setting. For example, algorithms developed on data from various acquisition settings may be predominantly applied in hospitals with lower quality data acquisition, leading to an inadvertent performance drop. We formulate subgroup shift detection in the framework of statistical hypothesis testing and show that recent state-of-the-art statistical tests can be effectively applied to subgroup shift detection on medical imaging data. We provide synthetic experiments as well as extensive evaluation on clinically meaningful subgroup shifts on histopathology as well as retinal fundus images. We conclude that classifier-based subgroup shift detection tests could be a particularly useful tool for post-market surveillance of deployed ML systems.\footnote{Find our code at \url{https://github.com/lmkoch/dht-subgroup}}

\end{abstract}


\section{Introduction}
\label{sec:introduction}


Machine learning (ML) tools for automated medical image interpretation have been approaching expert-level performance in controlled settings in recent years \cite{liu2019ComparisonDeepLearning}. However, major hurdles still obstruct the wide adoption of machine learning in clinical practice. When ML is applied in a clinical setting, its outputs are typically used to inform treatment decisions. Therefore, as a flipside to their vast potential, ML algorithms can also indirectly cause harm to the patient. For example, if such techniques fail to detect the presence of a tumour, patients may be treated inappropriately and in the worst case, even die prematurely. Similarly, in a screening setting, if ML algorithms perform poorly in detecting early stages of a disease, care for individual patients may be inferior, leading to increased burden on public health systems overall. 

Therefore, ML systems for healthcare are considered medical devices, and strict regulations (e.g. by the recently enacted Medical Device Regulation (MDR) in the European Union~\cite{mdr2017}) facilitate their safety and effectiveness. Regulatory approval processes for ML systems are still being shaped, e.g. with the draft European AI Act~\cite{eu2021draftai}, or with the AI/ML-based Software as a Medical Device Action Plan by the U.S. Food and Drug Administration~\cite{FDA2021}. The medical AI research community is contributing to this process, for example with a recently proposed medical algorithmic audit framework~\cite{LIU2022e384}.

\begin{figure*}
\centerline{\includegraphics{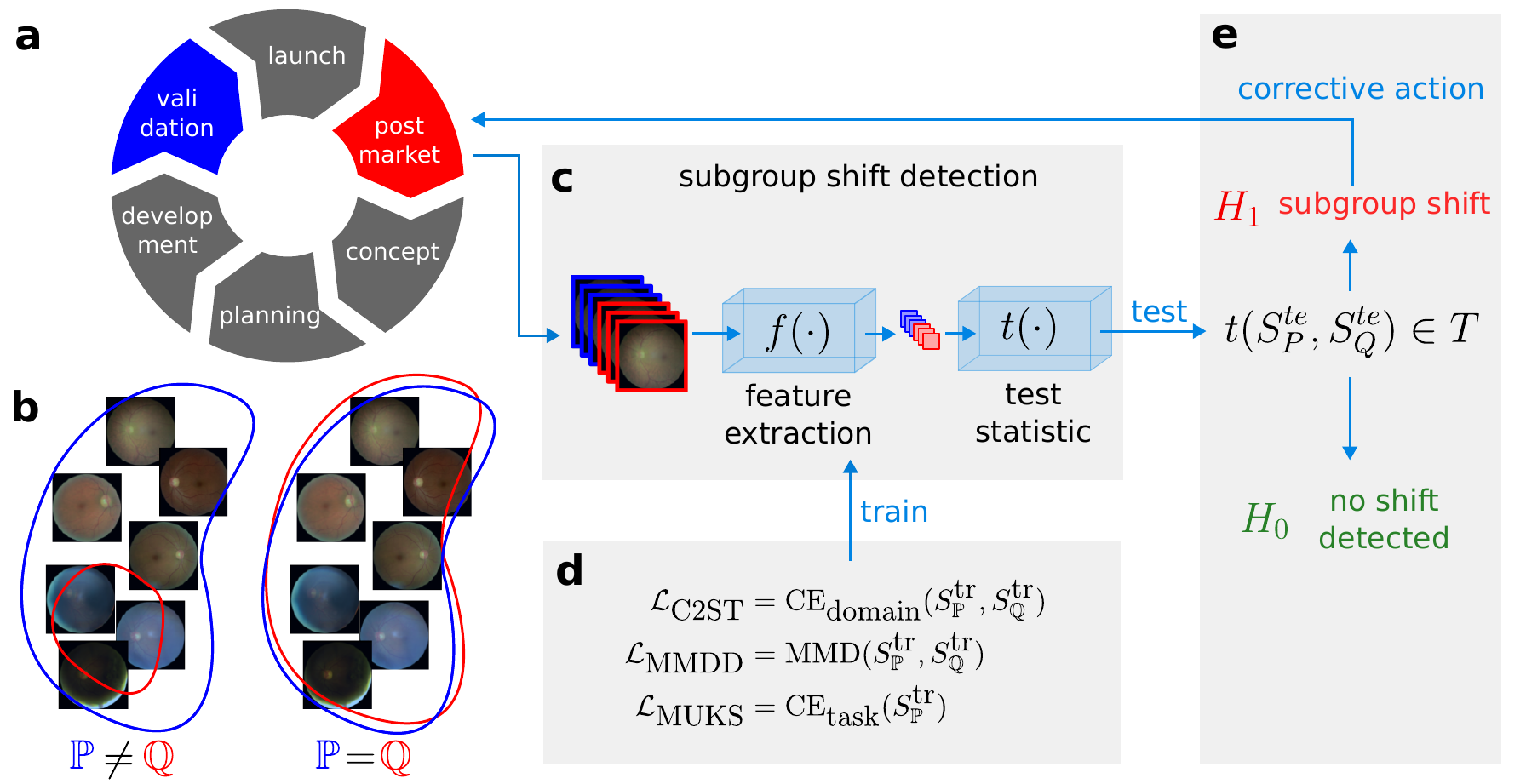}}
\caption{Overview of the proposed subgroup shift detection framework in the context of postmarket surveillance of a medical image-based ML device: \textbf{(a)} the medical device lifecycle including validation (blue) and postmarket (red) phases, \textbf{(b)} illustration of potential subgroup shifts between the source distribution $\mathbb{P}$ (blue) and the target $\mathbb{Q}$ (red), where all images from  $\mathbb{Q}$ are within the support of the source distribution and thus cannot be detected as outliers. \textbf{(c)} Subgroup shifts can be detected using neural network-based hypothesis tests consisting of a deep feature extractor and a test statistic. \textbf{(d)} The feature extractors and their optimisation depend on the shift detection approach. \textbf{(e)} Finally, a hypothesis test is carried out on held-out test samples. If a subgroup shift is detected, its causes can be investigated and corrective action can be taken in the context of post-market activities of a medical device manufacturer.}
\label{fig:graphical-abstract}
\end{figure*}

Such medical device regulations govern the whole life cycle of a product (Fig.~\ref{fig:graphical-abstract}\,\textbf{a}). One key step in this process is clinical validation, where the performance is assessed on data that are intended to be representative of the real data distribution encountered by the deployed system. After clinical validation, the system may be certified as safe and approved for use in the intended setting. ``Safe'' in this context means that the benefits of using the ML system are considered to outweigh the risks associated with prediction errors. This risk-benefit analysis crucially hinges on realistic ML performance estimates: if the performance in the deployment setting falls short of the claimed performance estimated in the validation setting, the cumulative harm associated with prediction errors in all patients may exceed the level deemed acceptable, and specifically, the level reported during validation. To demonstrate that a medical device stays within the validated performance specifications during real-life use, medical device manufacturers are required to implement post-market surveillance (PMS) measures that continuously monitor the deployed system during the postmarket phase (highlighted in red in Fig.~\ref{fig:graphical-abstract}\,\textbf{a}). The requirements for PMS activities have become stricter with the recent introduction of the MDR\,\cite{mdr2017} and the FDA's medical device action plan~\cite{FDA2021}. They are of particular importance for opaque deep learning systems, where failure mechanisms are difficult to inspect and changes over time in ML systems' input data and outputs  may therefore go unnoticed. 

In reality, changes in performance can be caused by a myriad of factors that induce data distribution shifts with respect to the validation setting. Examples include changes in technology (for example affecting low-level characteristics of acquired images), changes in patient demographics or changes in behaviour (for example increased screening for a disease due to new guidelines) \cite{finlayson2021ClinicianDatasetShift}. Many of these changes can be subtle and difficult to detect by humans, but may lead to unexpected changes in performance \cite{LIU2022e384}.
%
%

One type of dataset shift that is particularly difficult to detect is subgroup shift. Subgroup shifts can occur when the prevalence of individual subgroups is different in real-world data encountered by the deployed algorithm compared to the clinical validation data. 
For example, an algorithm for screening for ophthalmic diseases may be evaluated on a heterogeneous dataset spanning multiple sites, various imaging protocols and devices, and including a diverse patient population. While validation on heterogeneous data is in principle highly desirable, there may be (hidden) stratification in the data, and performance in subgroups may vary distinctly \cite{oakden2020hiddenStratification,eyuboglu2022domino}. It is possible that the ML system is then predominantly applied to a subgroup of the original population: this can happen if the ML system is deployed to screen patients in hospitals with lower-quality imaging procedures or hospitals with a catchment area predominantly inhabited by members of certain ethnicities. Such factors lead to a distribution shift which can impact the real-world performance, or will at least render original performance claims invalid. Both scenarios pose serious problems for safely deploying machine learning systems.
While a shift could be detected by measuring and monitoring the distribution of patient attributes and acquisition settings, crucial covariates characterising relevant subgroup attributes are often unmeasured or unidentified \cite{oakden2020hiddenStratification,eyuboglu2022domino}. 

Since subgroups are within the support of the original distribution, no individual data point of a subgroup is atypical or ``out-of-distribution'' (OOD). OOD detection is therefore unsuitable for detecting subgroup shifts~\cite{koch2022hidden}.

In this paper, we address the problem of detecting clinically relevant subgroup shifts in medical imaging data, and argue that our solution can be employed as part of the post-market surveillance strategy of a medical device manufacturer. To detect such shifts, we propose to use the framework of statistical hypothesis testing with a null hypothesis that two sets of datapoints are drawn from the same distribution. Recently developed hypothesis tests have reached meaningful statistical power on high-dimensional data \cite{liuLearningDeepKernels,rabanser2019FailingLoudlyEmpirical,cheng2019classification}, but this problem has so far not been explored in a medical imaging setting. We focus on the following key contributions: 

\begin{enumerate}
    \item We show that the performance in a range of medical image classification tasks varies between subgroups, motivating the need for detecting potential subgroup distribution shifts.
    \item We provide an overview of hypothesis testing approaches that are applicable to subgroup distribution shift detection in  high-dimensional imaging data. 
    \item We establish the state-of-the-art for detecting  various subgroup shifts  on toy data, retinal fundus images and histopathology images. 
\end{enumerate}

This paper is an extension of our previous proof of concept \cite{koch2022hidden}, covering additional subgroup shift detection techniques,  extensive experimental evaluation of subgroup performance disparities and subgroup shift detection in clinically relevant and realistic settings.



\section{Related Work}

Different approaches exist to mitigate problems arising from subgroup shifts and potential subgroup performance disparities. The first solution approach is to train algorithms to perform robustly across subgroups. While this does not prevent distribution shifts from occurring in the first place, robust performance across subgroups could prevent performance drop in the deployed system. Such distributionally robust algorithms have for example been presented in \cite{shimodaira2000improving,Sagawa2020Distributionally,cui2019classbalanced}, and a thorough overview and benchmark is provided in~\cite{wilds2021}. One important limitation of these approaches is that typically, relevant subgroup attributes must be known apriori and measured. 

Another body of literature aims to discover previously unknown meaningful subpopulations in the data. For example, \cite{oakden2020hiddenStratification} proposed strategies to identify hidden stratifications, but relied on experts for finding clinically relevant attributes. Alternatively, a low-dimensional embedding can reveal subgroups with systematic performance disparities, where subgroups are automatically labelled with textual descriptors  \cite{eyuboglu2022domino,jain2023distilling}. These approaches could later enable monitoring relevant subgroups during model deployment, and could lead to new insights during model development. 

Thus, related work so far either relied on subgroup labels for training robust models, or aimed at identifying relevant stratifications as a preliminary step which could facilitate downstream intervention. In contrast, our proposed approach for subgroup shift detection is agnostic of metadata and operates directly and solely on the input images by detecting whether two groups of images are systematically different. We argue that subgroup shifts can be detected as distribution shifts at deployment stage. While not specifically aimed at this application, several distribution shift detection techniques have recently been proposed \cite{kubler2022automl,rabanser2019FailingLoudlyEmpirical,liuLearningDeepKernels,cheng2019classification,lopez2016revisiting}. These methods are essentially neural-network based hypothesis tests, and have so far achieved reasonable test power in variations of MNIST and toy natural image datasets such as CIFAR-10.

The distribution shift detection methods introduced above operate on a group level. In contrast to these, there exists a large body of traditional OOD methods (see e.g. \cite{yang2022openood} for a comprehensive overview). These aim to discover individual OOD data points and, as we have shown in \cite{koch2022hidden}, cannot be applied to detect subgroup shifts.

\section{Subgroup Shift Detection Framework}
\label{sec:methods}

We present a general monitoring framework which can serve as a tool during the postmarket surveillance of deployed medical AI algorithms (see Fig.\,\ref{fig:graphical-abstract}). Our goal is to detect distribution shifts between the validation and postmarket (i.e. deployment) phases of a medical AI device (Fig.\,\ref{fig:graphical-abstract}\,\textbf{a}). We exclude the more widely studied OOD shift detection from the scope of this work, as a large body of work already exists for that setting \cite{yang2022openood}. Instead,  we specifically focus on the difficult task of detecting subgroup shifts (Fig.\,\ref{fig:graphical-abstract}\,\textbf{b}), where the target distribution $\mathbb{Q}$ comprises a subgroup within the domain of the source distribution $\mathbb{P}$.

\subsection{Framework Overview}

We assume a clinically validated image-based ML algorithm, and study the scenario where the manufacturer (or an auditor) has access to data from the validation and the real-world postmarket phase. We refer to these source and target distributions as $\mathbb{P}$ and $\mathbb{Q}$ over $\mathcal{X}$ and $\mathcal{Y}$, respectively. $\mathcal{X} \subseteq \mathbb{R}^n, \mathcal{Y} \subseteq \mathbb{R}^n$ are subsets of the space of $n$-dimensional images. We formally define a subgroup shift as the scenario where $\mathcal{Y} \subseteq  \mathcal{X}$ and $\mathbb{P} \neq \mathbb{Q}$. Following \cite{casella2002statistical,gretton2012KernelTwoSampleTest}, we formulate the problem of subgroup shift detection as a hypothesis test with null hypothesis $H_0: \mathbb{P} = \mathbb{Q}$ and alternative $H_1: \mathbb{P} \neq \mathbb{Q}$. 
In statistical hypothesis testing, the hypotheses make a statement about a population parameter, and the test statistic $t(X, Y)$ is the corresponding estimate from the samples $X = \{x_i\}_{i=0}^m \stackrel{iid}{\sim} \mathbb{P}$ and $Y = \{y_i\}_{i=0}^m \stackrel{iid}{\sim} \mathbb{Q}$.
$H_0$ is rejected for some rejection region $T$ of $t$. The significance level $\alpha$ or Type I error denotes the probability that $H_0$ is rejected even if it is true. Typically, the rejection region is selected at a specific significance level, e.g. $\alpha = 0.05$. The test power denotes the probability that $H_0$ is correctly rejected if $H_1$ is true. Therefore, the test power is a useful performance measure for shift detection. 

Essentially, neural-network based hypothesis tests \cite{liu2019ComparisonDeepLearning,rabanser2019FailingLoudlyEmpirical,cheng2019classification} rely on feature representations of the high-dimensional image data in which the two domains $\mathbb{P}$ and $\mathbb{Q}$ are well separated. The procedure for applying such a test consists of two steps: We first fit the parameters of the neural-network based feature extractors on a training fold $S_{\mathbb{P}}^{\mbox{tr}}, S_{\mathbb{Q}}^{\mbox{tr}}$ of both the source and target distributions (Fig.\,\ref{fig:graphical-abstract}\,\textbf{c, d}),  and then perform a hypothesis test on the deep features of the test fold $S_{\mathbb{P}}^{\mbox{te}}, S_{\mathbb{Q}}^{\mbox{te}}$ from the source and target data (Fig.\,\ref{fig:graphical-abstract}\,\textbf{e}). If a subgroup shift is detected, this would indicate that the deployment setting of the ML algorithm may violate its intended use. An alert could be issued and corrective action could be taken by the manufacturer to investigate the reason and implications of the observed shift (Fig.\,\ref{fig:graphical-abstract}\,\textbf{e}).
 
Based on our own prior work\cite{koch2022hidden} and a performance comparison on toy data in \cite{liu2019ComparisonDeepLearning,rabanser2019FailingLoudlyEmpirical}, we have identified three hypothesis testing frameworks as suitable candidates for detecting subgroup distribution shifts in medical images. These are described in detail in the following sections.

\subsubsection{Classifier-Based Tests (C2ST)}
\label{sec:ct}

Classifier-based hypothesis tests make use of a domain classifier that aims to discriminate between examples from $\mathbb{P}$ and $\mathbb{Q}$ \cite{lopez2016revisiting,cheng2019classification}. As domain classification becomes impossible for $\mathbb{P} = \mathbb{Q}$, this motivates using a measure based on the classification performance as a test statistic $t(X, Y)$. The null hypothesis could then be rejected in favour of $H_1: \mathbb{P} \neq \mathbb{Q}$ for significantly above chance performance. \cite{cheng2019classification} proposed a test statistic based on the classifier's logit output $f_{\mbox{C2ST}}: \mathcal{X} \to \mathbb{R}$:
\begin{align}
    t(X,Y) &= \frac{1}{m} \sum_{i} f_{\mbox{C2ST}}(x_i) - \frac{1}{m} \sum_{j} f_{\mbox{C2ST}}(y_j)
\label{eq:test-statistic-c2st}
\end{align}
Thus, the domain classifier can be viewed as a feature extractor (Fig.\,\ref{fig:graphical-abstract}\,\textbf{c}) whose parameters are fit on training images from the source and target distribution $S_{\mathbb{P}}^{\mbox{tr}}, S_{\mathbb{Q}}^{\mbox{tr}}$, using for example a cross-entropy loss to distinguish examples from $\mathbb{P}$ and $\mathbb{Q}$ (Fig.\,\ref{fig:graphical-abstract}\,\textbf{d}). Fitting the parameters of the C2ST feature extractor therefore requires only knowledge of the domain membership, but no labels with regard to a specific task.
The test statistic from Eq.\,\ref{eq:test-statistic-c2st} can be calculated on two held out samples from the source and training data $S_{\mathbb{P}}^{\mbox{te}}, S_{\mathbb{Q}}^{\mbox{te}}$ and a permutation test can be carried out to determine the rejection threshold (Fig.\,\ref{fig:graphical-abstract}\,\textbf{e}).

\subsubsection{Deep Kernel Tests (MMDD)}
\label{method:mmd-d}

Deep kernel tests \cite{liuLearningDeepKernels} are a recent generalisation of kernel two-sample tests \cite{gretton2012KernelTwoSampleTest}, which use a distance metric between probability distributions as a test statistic to examine whether $H_0: \mathbb{P} = \mathbb{Q}$. Intuitively, a suitable distance metric should be low if the null hypothesis is true (i.e. the distributions $\mathbb{P}, \mathbb{Q}$ are the same), and higher for $\mathbb{P} \neq \mathbb{Q}$, thus allowing us to reliably reject $H_0$. One such a distance metric is the Maximum Mean Discrepancy (MMD) on a Reproducing Kernel Hilbert Space (RKHS)\cite{gretton2012KernelTwoSampleTest}:

\begin{definition}
\cite{gretton2012KernelTwoSampleTest} Let $\mathcal{H}_k$ be a RKHS with kernel $k: \mathcal{X} \times \mathcal{X} \to \mathbb{R}$. The MMD is defined as

\begin{equation}
    \mbox{MMD}[\mathcal{H}_k, \mathbb{P}, \mathbb{Q}] = \sup_{f \in \mathcal{H}_k, \norm{f}_{\mathcal{H}_k} \le 1} \left( \mathbb{E}_{x \sim \mathbb{P}}[f(x)] - \mathbb{E}_{y \sim \mathbb{Q}}[f(y)] \right) ,
\end{equation}
and an unbiased estimator for the MMD is
\begin{align}
    \widehat{\mbox{MMD}}(X, Y; k) & = \frac{1}{m(m-1)} \sum_{i \neq j} H_{ij} \quad , \label{eq:mmd} \\
    H_{ij} & = k(x_i, x_j) + k(y_i, y_j) - k(x_i, y_j) - k(y_i, x_j)
\end{align}
\end{definition}

For characteristic kernels $k$, the MMD is a metric, which implies that $\mbox{MMD}[\mathcal{H}_k, \mathbb{P}, \mathbb{Q}] = 0$ iff $\mathbb{P} = \mathbb{Q}$.
The metric property makes $t(X,Y) = \widehat{\mbox{MMD}}(X, Y)$ an appropriate test statistic for testing whether $\mathbb{P} = \mathbb{Q}$. 

The choice of the kernel $k$ affects the test power in finite sample sizes and developing suitable kernels is an active area of research (e.g. \cite{sutherland2021GenerativeModelsModel,liuLearningDeepKernels}).
We follow \cite{liuLearningDeepKernels} and parameterise the kernel $k$ with a neural network. Specifically, \cite{liuLearningDeepKernels} use a neural network $f_{\mbox{MMDD}}: \mathcal{X} \to \mathbb{R}^{128}$ as a feature extractor and define the final kernel as a combination between two Gaussians kernels operating on the original image space and the feature space, respectively:
\begin{equation}
k(x, y) = \left( (1-\delta) g_a(f(x), f(y)) + \delta \right) g_b(x, y) \quad ,
\label{eq:kernel}
\end{equation}
where $g_a, g_b$ are Gaussians with length scales $\sigma_a, \sigma_b$. The kernel parameters, including neural network parameters $\theta$, are optimised on training images from $\mathbb{P}$ and $\mathbb{Q}$ by maximising the Maximum Mean Discrepancy:
%
\begin{align}
    \mathcal{L}_{\mbox{MMDD}} &= - \widehat{\mbox{MMD}}(S_{\mathbb{P}}^{\mbox{tr}},S_{\mathbb{Q}}^{\mbox{tr}}; k) \label{eq:mmd-objective}
\end{align}
We diverge slightly from \cite{liuLearningDeepKernels}, where the objective function also incorporated knowledge on the asymptotic distribution of $\widehat{\mbox{MMD}}$ under $H_1$. We did not find this beneficial, as reported in our prior work \cite{koch2022hidden}.
For a trained kernel, the test statistic $t(X,Y) = \widehat{\mbox{MMD}}(X, Y)$ can be calculated on samples $X,Y$ from the test fold of the source and target data $S_{\mathbb{P}}^{\mbox{te}}, S_{\mathbb{Q}}^{\mbox{te}}$, and again a permutation test can be used to determine whether  $H_0: \mathbb{P} = \mathbb{Q}$ can be rejected.

\subsubsection{Multiple Univariate Kolmogorov-Smirnov Tests on Task Predictions (MUKS)}
\label{method:rabanser}

Finally, \cite{rabanser2019FailingLoudlyEmpirical} propose hypothesis tests for domain shift detection that operate on low-dimensional representations of the original image space $\mathcal{X}$. They use a black-box task classifier $f_{\mbox{MUKS}}: \mathcal{X} \to \mathbb{R}^C$ as a dimensionality reduction technique, where $s = f(x) \in \mathbb{R}^C$ is a softmax prediction for $x$ with $C$ classes. Essentially, in our case $f_{\mbox{MUKS}}$ can be the monitored ML algorithm itself. For simplicity, we train $f_{\mbox{MUKS}}$ using images from $\mathbb{P}$ and associated labels.

Multiple univariate Kolmogorov-Smirnov (MUKS) test are then applied to the softmax predictions of samples $X$ and $Y$ from the test fold $S_{\mathbb{P}}^{\mbox{te}}, S_{\mathbb{Q}}^{\mbox{te}}$. 
The two-sample KS test statistic can be calculated for each class individually as 
\begin{align}
t_c(X,Y) &= \sup_{f_c(x)} |F_{X,c}( f_c(x))-F_{Y,c}(f_c(x))| ~ ,
\label{eq:test-statistic-muks}
\end{align}
where $F_{X,c}, F_{Y,c}$ are the empirical distribution functions over the softmax outputs of samples $X, Y$. Standard KS tests can then be carried out, yielding $C$ $p$-values. As $C$ tests are performed, the $p$-values must be corrected for multiple comparisons. As in
\cite{rabanser2019FailingLoudlyEmpirical}, we perform Bonferroni correction, i.e. $H_0$ is rejected if $p \le \alpha / C$ for any of the $C$ comparisons.

\subsubsection{Required data}

\begin{table}[b]
\centering
\caption{Data requirements for the compared distribution shift detection methods.}
\label{tab:method-train-data-reqs}
\begin{tabular}{@{\extracolsep{4pt}}lllll@{}}
\hline
\multirow{2}{*}{\textbf{Method}} & \multicolumn{2}{c}{\textbf{Train}} & \multicolumn{2}{c}{\textbf{Test}} \\ \cline{2-3} \cline{4-5}
 & \textbf{Images} & \textbf{Labels} & \textbf{Images} & \textbf{Labels} \\
\hline
C2ST & $\mathbb{P}, \mathbb{Q}$ & - & $\mathbb{P}, \mathbb{Q}$ & - \\
MMDD & $\mathbb{P}, \mathbb{Q}$ & - & $\mathbb{P}, \mathbb{Q}$ & - \\
MUKS & $\mathbb{P}$ & $\mathbb{P}$ & $\mathbb{P}, \mathbb{Q}$ & - \\
\hline
\end{tabular}
\end{table}

The deep hypothesis tests used for subgroup shift in this paper use training data for learning feature extractors to compute test statistics, and then perform the hypothesis tests on test data (see Table\,\ref{tab:method-train-data-reqs}). The first two methods, C2ST and MMDD, require training images from both source and target distributions without additional labels. In contrast, MUKS, the final studied method, does not require access to the target distribution for training. Instead, it uses images from the source distribution with ground truth labels for the monitored ML algorithm's task. None of the methods require task labels from the target distribution. While the real-world data availability may depend on the specific application and deployment route of a medical device, we argue that all three methods can be applicable in postmarket surveillance procedures, where no task labels can be expected, but access to production data should be available.

\section{Experiments and Results}

We evaluated the above shift detection methods in a variety of synthetic (Sec. \ref{sec:data-mnist}) and real medical image datasets including histopathology images (Sec. \ref{sec:data-camelyon}) and retinal fundus images (Sec. \ref{sec:data-eyepacs}). We will make the code publicly available upon publication. For each dataset, we first analysed classification accuracy for an associated task (e.g. Diabetic Retinopathy grading for retinal fundus images) on the whole test set as well as in the studied subgroups. We aimed to achieve classification accuracy approximately in line with the current state-of-the-art but emphasise that our main goal was to investigate  subgroup performance disparities. The potential for subgroup disparities then motivated the accurate detection of subgroup shifts.  

In the central experiments of this paper, we modelled a variety of subgroup shifts caused by changes in acquisition settings, demographics or medical history of the patients. We then assessed the shift detection performance of the methods presented in Sec.~\ref{sec:methods} on all studied datasets. Importantly, our shift models and evaluation required knowledge of the subgroup attributes. However, in real-world settings, access to relevant attributes is often not available, and the subgroup shift detection methods used here do not rely on them.

\subsection{Synthetic example}
\label{sec:proof-of-concept}

\begin{figure}
\centerline{\includegraphics{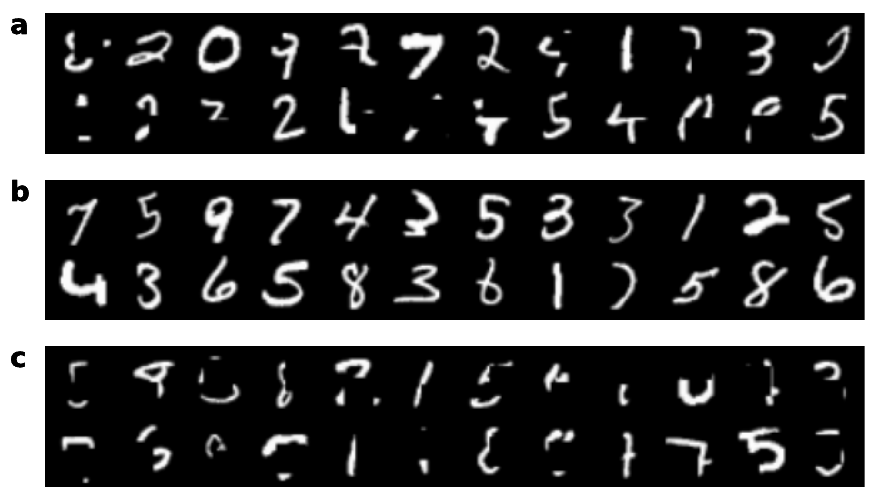}}
\caption{\textbf{a}) Modified MNIST data with hidden stratification:  \textbf{b}) a subgroup of the digits were intact and \textbf{c}) a subgroup were artificially corrupted, substantially occluding the digits .}
\label{fig:mnist-hidden-subgroups}
\end{figure}

We first demonstrated the problem setting on a synthetic example, where we simulated hidden subgroups in the MNIST dataset by altering parts of the data, thereby making digit classification systematically more difficult for one subgroup. 

\subsubsection{Data}
\label{sec:data-mnist}

We used the MNIST dataset \cite{lecun1998mnist} with the official training and test splits (60'000 / 10'000 images). We held out 10'000 images from the training split for hyperparameter and model selection. We modelled hidden subgroups by randomly adding rectangular occlusions to the images in $\mathbb{P}$ with a probability of $p=0.5$ (see Fig.~\ref{fig:mnist-hidden-subgroups}), such that the training data consisted of two subgroups, one with normal digits and one with occluded digits. A subgroup distribution shift was then modelled by adjusting the proportion of occluded images in $\mathbb{Q}$ to $p=1$ (corrupted images only, Fig.\,\ref{fig:mnist-hidden-subgroups}\,\textbf{c}).

\subsubsection{Digit classification on corrupted MNIST}

We trained a digit classifier on the partially corrupted MNIST data using a ResNet-18 architecture and a cross-entropy loss. The task classifier $f_{\mbox{MUKS}}$ was trained on the training set consisting of both subgroups of intact and corrupt digits.

The overall classification accuracy on the test set was $0.952$ (Table\,\ref{tab:mnist-subgroup-results}). When the subgroups were evaluated separately using the same model, the performance varied, with the corrupted images yielding a worse accuracy of $0.919$. In this case, an undetected shift in the deployed system would be harmful, motivating the detection of potential subgroup shifts.

\begin{table}[b]
\centering
\caption{Subgroup analysis of task performance: digit classification accuracy on MNIST.}
\label{tab:mnist-subgroup-results}
\begin{tabular}{llll}
\hline
\textbf{Attribute} & \textbf{Group} & \textbf{n} & \textbf{Acc.} \\
\hline
All       &       & 10000 & 0.952 \\
\hline
\multirow{2}{*}{Corruption}
  & All intact & 10000  & 0.986 \\
  & All corrupt & 10000   & 0.919 \\
\hline
\end{tabular}
\end{table}

\subsubsection{Deep hypothesis tests reliably detect subgroup shifts}
\label{sec:exp-mnist-subgroup-shift}
\label{sec:exp-mnist}

We next studied the performance in detecting a subgroup shift $\mathbb{P} \rightarrow \mathbb{Q}$ using the techniques proposed in Sec~\ref{sec:methods}. We considered the problematic subgroup of corrupted digits (Fig.~\ref{fig:mnist-hidden-subgroups}\,c) as distribution $\mathbb{Q}$. The neural networks forming the backbones of the respective hypothesis tests for shift detection $f_{\mbox{MMDD}}, f_{\mbox{C2ST}}, f_{\mbox{MUKS}}$ were trained using training and validation folds of data from both $\mathbb{P}$ and $\mathbb{Q}$. We then assessed the shift detection rate of each method through the test power, i.e. the proportion of rejected null hypotheses $H_0: \mathbb{P} = \mathbb{Q}$ in repeated experiments ($100$ repetitions at significance level $\alpha=0.05$). For each repetition, we drew samples with replacement of size $m \in \{10, 30, 50, 100, 200, 500\}$ from the test folds of $\mathbb{P}$ and $\mathbb{Q}$. We calculated the empirical Type I error similarly by repeatedly drawing two independent samples of $\mathbb{P}$.

For the classifier-based test C2ST, we trained a domain classifier to discriminate between $\mathbb{P}$ and $\mathbb{Q}$ using the same configuration as for the task classifier, i.e. using a ResNet-18 backbone. 
For the deep kernel test MMDD, we trained a kernel to maximise the MMD between $\mathbb{P}$ (all images as in Fig.~\ref{fig:mnist-hidden-subgroups}\,a) and $\mathbb{Q}$ by optimising the objective from Eq.\,\ref{eq:mmd-objective} on the training fold. The feature extractor $f_{\mbox{MMDD}}(x)$ was chosen as in \cite{liuLearningDeepKernels} and consisted of a shallow convolutional neural network with four convolutional layers and a linear layer with $128$ outputs. We used the Adam optimiser \cite{kingma2015adam} with a learning rate of $10^{-4}$. 
For the Multiple Univariate Kolmogorov-Smirnov Test (MUKS), we reused the task classifier from the analysis of subgroup performance disparities above, and applied MUKS to softmax predictions of test samples from $\mathbb{P}$ and $\mathbb{Q}$.
For all methods, we used a batch size of 64 and selected the models with the best loss on the validation set.

\begin{figure}
\centering{\includegraphics{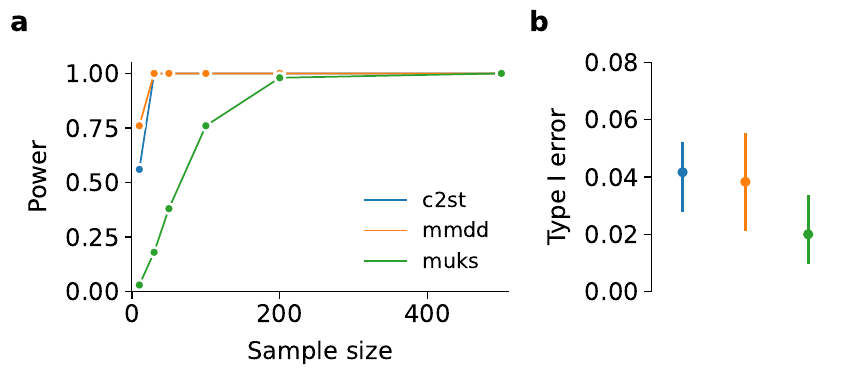}}
\caption{\textbf{a)} Detection performance for subgroup shifts in corrupted MNIST, \textbf{b)} Type I error for each detection method.}
\label{fig:mnist-results}
\end{figure}

Both C2ST and MMDD were able to detect shifts reliably with a sample size of at least 100, with C2ST markedly outperforming MMDD (Fig.\,\ref{fig:mnist-results}\,\textbf{a}). While MUKS power increased with increasing sample size, it stayed low even for $m=500$. As expected, for all tests, the type I error (Fig.~\ref{fig:mnist-results}\,\textbf{b}) remained close to the chosen significant level $\alpha=0.05$. The type I error for MUKS was notably lower, which we attributed to the overly conservative Bonferroni correction.

\subsection{Proof of concept in histopathology images}

Next, we performed experiments on publicly available histopathology images, which vary severely across sites due to differences in sample preparation and staining protocols. The setting is therefore popular for evaluating domain generalisation research\,\cite{wilds2021}, and we use it here to model subgroup shifts in acquisition settings by changing the prevalence of individual hospitals in the target distribution. 

\subsubsection{Data}
\label{sec:data-camelyon}

\begin{figure}
\centerline{\includegraphics{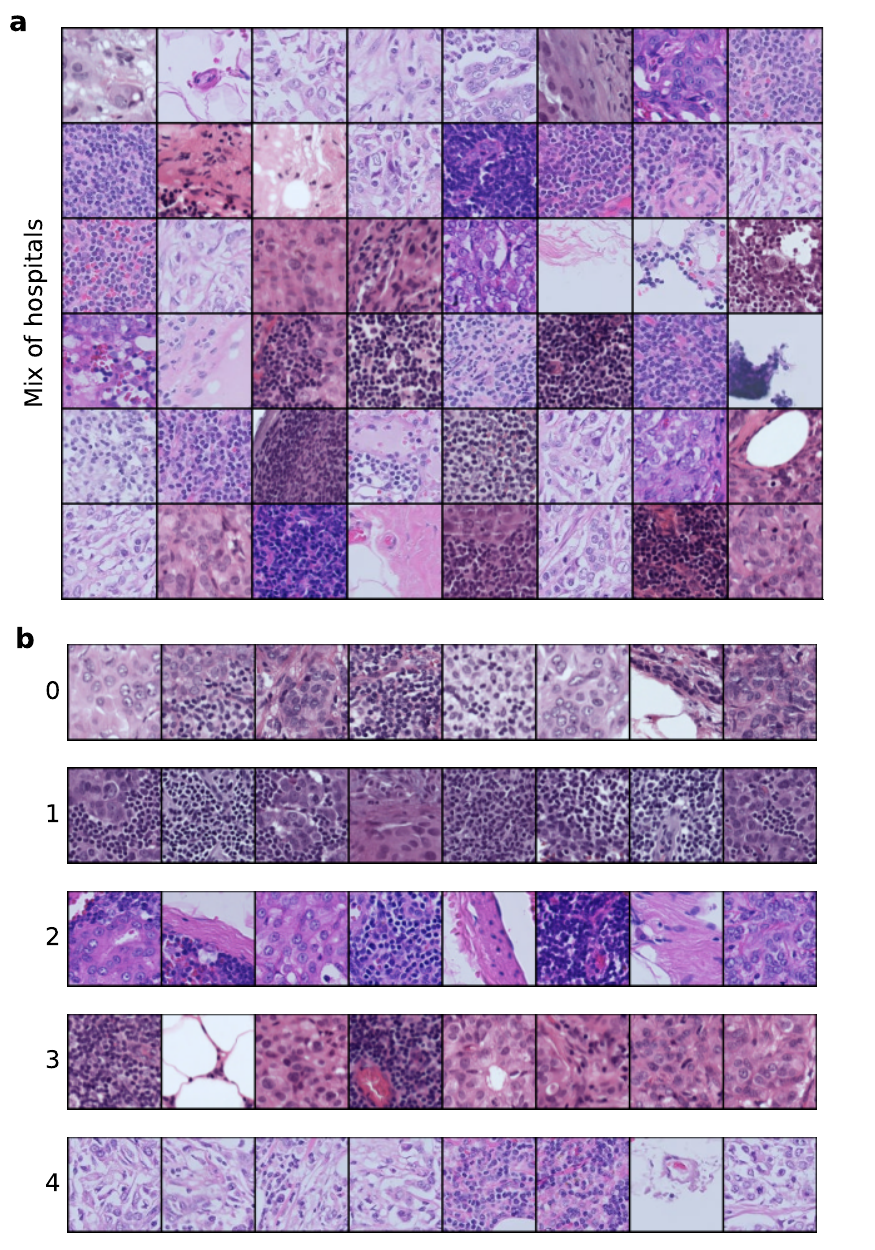}}
\caption{\textbf{a)} Example patches from the heterogeneous dataset including all hospitals, \textbf{b)} data from individual hospitals 0--4 reveals systematic differences across acquisition sites.}
\label{fig:camelyon-example-data}
\end{figure}

We used the Camelyon17 challenge data \cite{bandi2019camelyon}, which consisted of 50 whole slide images (WSI) of H\&E stained lymph node biopsies acquired across $5$ different hospitals (see examples in Fig.~\ref{fig:camelyon-example-data}).  We used a patch-based version of Camelyon17 \cite{wilds2021} to study tumor detection across different hospitals. 
We split the data into training (284'219), validation (70'219) and test patches (100'820) of size 96x96 while making sure no data from the test slides were used in the training and validation folds. Tumour prevalence was approximately $50\%$ in all folds. Subgroup shifts were modelled by including only data from a single site in the target distribution $\mathbb{Q}$.

\subsubsection{Tumor detection performance in subgroups}
\label{sec:exp-camelyon-subgroup-performance}

We trained a ResNet-50 model for the detection of tumor cells on the training fold of the source data $\mathbb{P}$ using a cross-entropy loss. The loss was optimised using stochastic gradient descent with Nesterov momentum, an initial learning rate (LR) of 0.001 and LR decay of 0.9. Data augmentation included random cropping, horizontal and vertical flipping, color distortion and rotation. The model was trained for 25 epochs, and the final model was chosen based on the validation loss.

The overall accuracy on the test fold of Camelyon17 was $0.860$ but varied considerably between $0.759$ and $0.926$ between different hospitals (see Table~\ref{tab:camelyon-subgroup-results}). Accuracy did not seem to depend on the individual subset sizes, i.e., all groups were adequately represented. Again, the subgroup performance disparities suggested that detecting potential hospital shifts could be important in a deployment setting.

\begin{table}[b]
\centering
\caption{Subgroup analysis of task performance: Tumor detection accuracy on Camelyon17.}
\label{tab:camelyon-subgroup-results}
\begin{tabular}{llll}
\hline
\textbf{Attribute} & \textbf{Group} & \textbf{n} & \textbf{Acc.} \\
\hline
All       &       & 100500 & 0.860 \\
\hline
\multirow{5}{*}{Hospital}
  & 0 & 12104  & 0.926 \\
  & 1 & 8412   & 0.830 \\
  & 2 & 16777  & 0.889 \\
  & 3 & 32243  & 0.925 \\
  & 4 & 30964  & 0.759 \\
\hline
\end{tabular}
\end{table}

\subsubsection{Classifier-based and kernel tests outperform MUKS for hospital subgroup shift detection}
\label{sec:exp-camelyon}

To evaluate the subgroup shift detection performance in histopathology data, we modelled a subgroup shift scenario where a tumor detection algorithm trained on a mix of hospitals was deployed to a single hospital.
For MUKS, we reused the tumor detection classifier described above. For the classifier-based test C2ST we trained a ResNet-50 domain classifier $f_{\mbox{C2ST}}$ with the same training regime as the task classifier. For MMDD, we again used the feature extractor architecture  proposed in \cite{liuLearningDeepKernels}, and maximised the MMD estimator using Adam with a learning rate of $10^{-6}$. 

\begin{figure}
\centering{\includegraphics{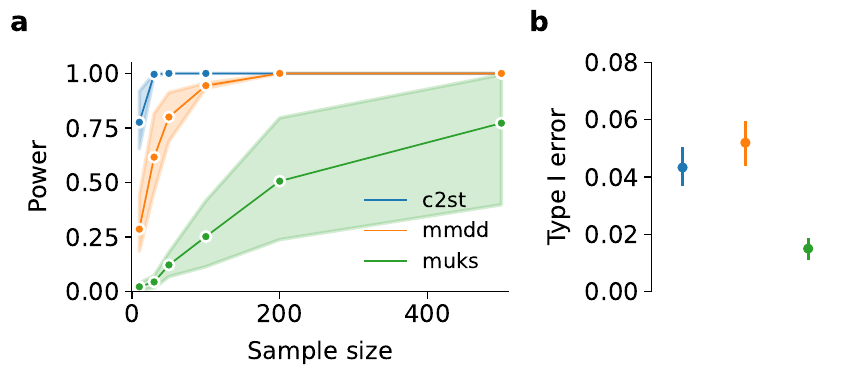}}
\caption{\textbf{a)} Shift detection performance for hospital subgroup shifts in Camelyon17 data, \textbf{b)} type I error. The shaded area denotes the variation ($95\%$ CI) between different hospitals.}
\label{fig:camelyon-results}
\end{figure}

The obtained results were consistent with the findings in MNIST: C2ST very reliably detected all hospital subgroup shifts, while MMDD also showed good performance for sample sizes from approximately $m=50$ (see Fig.~\ref{fig:camelyon-results}\,\textbf{a}). Test power dropped considerably for MUKS and there was a wider variance in performance depending on the individual hospital (shaded area in Fig.~\ref{fig:camelyon-results}\,\textbf{a}), again showing that MUKS tests performed less robustly in many settings. Similarly to the previous experiment, the Type I error was consistent with the significance level (see Fig.~\ref{fig:camelyon-results}\,\textbf{b}).

\subsection{Application to real-world retinal fundus images}
\label{sec:retinal}

Finally, we studied subgroup shifts in retinal fundus images that were acquired in the context of screening for Diabetic Retinopathy (DR). DR is a microvascular complication of diabetes that often leads to worsening symptoms of vision loss. It is a leading cause of blindness in the working age population and affects approximately 1 in 3 diabetes patients \cite{international2017ico}. Early intervention can slow disease progression \cite{international2017ico}, and diabetes patients therefore undergo regular screening. ML-assisted diagnosis of DR from retinal fundus images can therefore have a large public health impact globally. On the other hand, retinal fundus images vary widely in appearance, and image characteristics may be influenced by patient demographics \cite{mueller2022a} and acquisition details. Given the diversity and multitude of factors influencing the data and potentially an ML algorithms' performance, we argue that monitoring for distribution shifts can provide real benefit in a postmarket surveillance setting.

\subsubsection{Data}
\label{sec:data-eyepacs}

\begin{figure*}
\centerline{\includegraphics{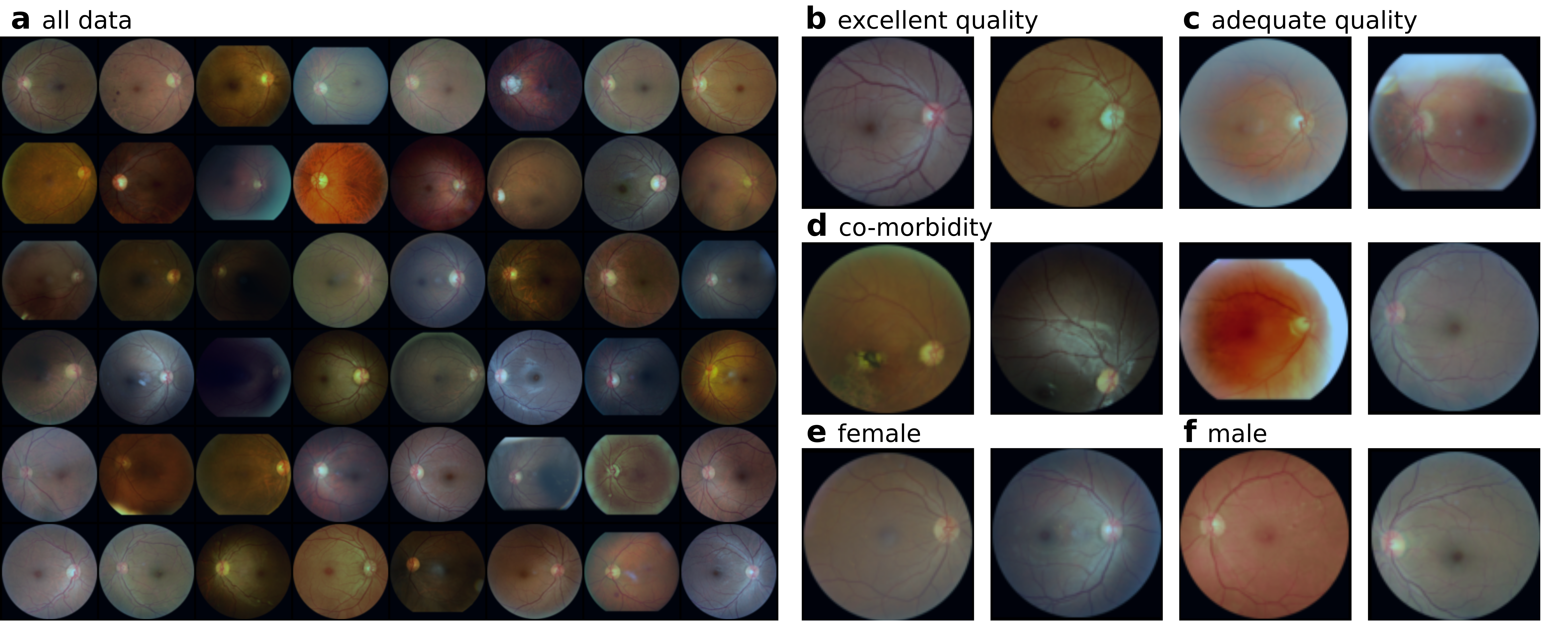}}
\caption{\textbf{a)} Example retinal fundus images from the Eyepacs dataset. Magnified on the right are example \textbf{(b)}  images of excellent and \textbf{(c)} adequate quality, \textbf{(d)} images with co-morbidities, and example images of \textbf{(e)} female  and \textbf{(f)} male patients .}
\label{fig:eyepacs-example-data}
\end{figure*}

We used a retinal fundus dataset from EyePACS, Inc., where a rich collection of relevant patient and image attributes were provided alongside the image data. The dataset consisted of 130'576 macula-centered images (see examples in Fig.~\ref{fig:eyepacs-example-data}\,\textbf{a}) which were manually assessed to be of at least adequate image quality for interpretation, and which contained self-reported information on patient sex and ethnicity. Furthermore, beyond expert DR grades, the presence of co-morbidities was reported including eye diseases such as cataract, glaucoma and others. 
Table~\ref{tab:eyepacs_description} provides a detailed overview of the included population with relevant patient and image attributes. Figures~\ref{fig:eyepacs-example-data}\,\textbf{b}-\textbf{f} highlights different image attributes and their impact on the retinal fundus images, including image quality (\textbf{b}-\textbf{c}), the presence of co-morbidities (\textbf{d}). In contrast, patient sex (\textbf{e}-\textbf{f}) cannot be visually inferred from the image by human experts.
We used this dataset to study potentially harmful subgroup shift scenarios where the target distribution $\mathbb{Q}$ may be consist of subgroups of the population, e.g. patients with existing co-morbidities, or data with lower image quality only.

The data was pre-processed by cropping a square bounding box fitting the circular fundus and resampling the images to a resolution of 512x512 pixels. The data were then split into training (78'126), validation (26'210) and test (26'150) folds.  

\begin{table}[]
\centering
\caption{Summary of relevant patient and image attributes in the EyePACS dataset, along with their prevalence in the training, validation and test folds. The rightmost column shows DR grading accuracy in subgroups of the test set. We did not report accuracy for subgroups with negligible sample size.}
\label{tab:eyepacs_description}
\begin{tabular}{llrrrl}
\hline
\multirow{2}{*}{Attribute} & \multirow{2}{*}{Group} & \multicolumn{3}{c}{Number of images} & \multirow{2}{*}{Acc.} \\ \cline{3-5}
\textbf{}      & \textbf{}           & \multicolumn{1}{l}{Training} & \multicolumn{1}{l}{Val.} & \multicolumn{1}{l}{Test} &  \\
\hline
All & \textbf{}           & 78126                        & 26210                         & 26150 & 0.868         \\
\hline
\multirow{3}{*}{Sex}            & Female & 46684 & 15795 & 15641 & 0.879 \\
 & Male & 31418 & 10413 & 10501 & 0.851 \\
 & Other & 24 & 2 & 8 & 0.875 \\
\hline
\multirow{7}{*}{Ethnicity}      & Latin American      & 54011 & 17908 & 18251 & 0.868                   \\
               & Caucasian           & 8648 & 2929 & 2977 & 0.900                   \\
               & African             & 6057 & 2006 & 1979 & 0.839                    \\
               & Asian               & 4362 & 1572 & 1273 & 0.895                   \\
               & Indian & 3675 & 1356 & 1212 & 0.810                   \\
               & Multi-racial        & 769 & 160 & 197 & 0.868
               \\
               & Native American     & 604 & 279 & 261 & 0.831 
               \\
\hline
\multirow{3}{*}{\shortstack{Image\\quality}}  & Excellent           & 14321 & 4892 & 4884 & 0.894 \\
               & Good & 32184 & 10712 & 10719 & 0.878 \\
               & Adequate            & 31621 & 10606 & 10547 & 0.845            \\
\hline
\multirow{2}{*}{\shortstack{Comorbi-\\dities}} & Present             & 7756 & 2698 & 2555 & 0.843 \\
               & Not present         & 70370 & 23512 & 23595 & 0.870 \\                 
\hline

\end{tabular}
\end{table}

\subsubsection{Subgroup performance disparities for Diabetic Retinopathy grading}

We first investigated the subgroup performance disparities of a DR grading algorithm to again analyse the potential for harm if subgroup shifts occur. We trained a ResNet-50 model that classified fundus images in 5 DR grades, ranging from 0 (healthy) to 4 (proliferative DR). We used a cross-entropy loss and the same training configuration as the tumour detection model on Camelyon17 (Sec.\,\ref{sec:exp-camelyon-subgroup-performance}). As before, this task classifier was trained on the training fold of $\mathbb{P}$, and then evaluated in individual subgroups.

The 5-class DR grading accuracy obtained in the EyePACS test set was $0.868$. Considerable performance differences were found across subgroups (right column in Table~\ref{tab:eyepacs_description}). For example, performance for different ethnicities ranged between $0.81$ and $0.90$, and classification accuracy was reduced for images with lower quality. Furthermore, we found a considerable accuracy drop for the group of patients with co-morbidities. A undetected distribution shift in the deployed ML system towards a higher prevalence of these patients could lead to increased patient harm.

\subsubsection{Classifier-based tests reliably detect subgroup shifts in retinal images}
\label{sec:exp-eyepacs}

\begin{figure*}
\centerline{\includegraphics{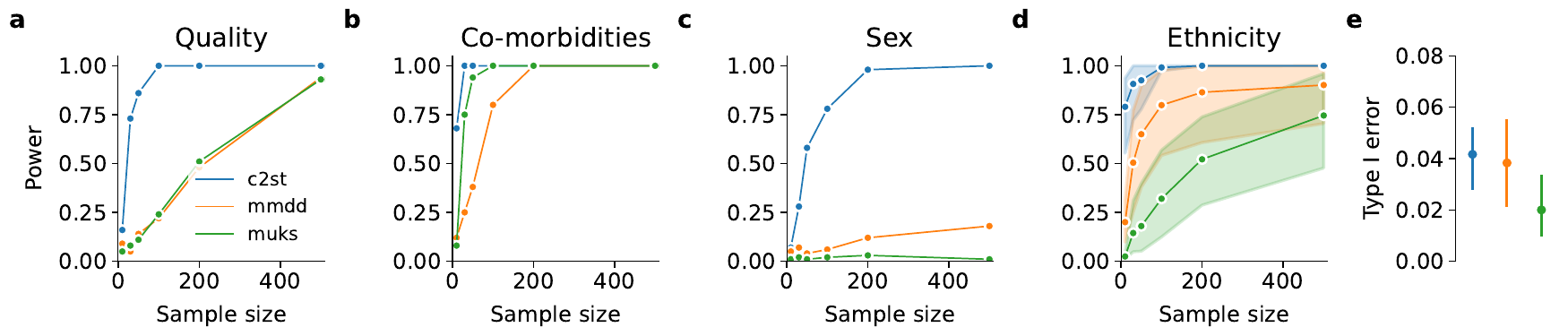}}
\caption{Shift detection results for subgroup shifts with respect to \textbf{(a)} image quality, \textbf{(b)} the presence of co-morbidities, \textbf{(c)} patient sex, and \textbf{(d)} patient ethnicity. \textbf{(e)} Shows the type I error for all methods evaluated on the Eyepacs dataset.}
\label{fig:eyepacs-main-results}
\end{figure*}

On retinal fundus images, we studied subgroup shifts with respect to patient sex, ethnicity, image quality and the presence of co-morbidities. For image quality, we simulated a subgroup shift by considering only ``Adequate'' images in the target distribution $\mathbb{Q}$, compared to all images of either adequate, good, or excellent quality in $\mathbb{P}$. To model subgroup shifts in the prevalence of co-morbidities, we included only the subgroup of images with co-morbidities in $\mathbb{Q}$. For both quality and co-morbidity shift, we observed considerable performance drops in these subgroups (see Tab.~\ref{tab:eyepacs_description}), making such distribution shifts potentially harmful if undetected. Furthermore, we argue that attributes such as image quality and presence of co-morbidities may not be routinely assessed in clinical validation settings and even less in deployment settings, and such shifts could thus go unnoticed unless detected based on the image distributions directly. We also modelled shifts in patient sex by including only images from female patients in the target distribution $\mathbb{Q}$, and ethnicity shifts by individually including only a single ethnicity in $\mathbb{Q}$. 


Similarly to the previous experiments, we trained neural networks for the hypothesis tests MMDD, C2ST and MUKS.
Again, we reused the task classifier as a feature extractor $f_{\mbox{MUKS}}$, and we trained a domain classifier $f_{\mbox{C2ST}}$ for the classifier-based test using a binary ResNet-50 and the same training configuration as the task classifier.
For the MMDD, some modifications were necessary compared to the previous experiments. Firstly, we reduced the image size from $512 \times 512$ to $96 \times 96$ as the original resolution was prohibitive for calculating the MMDD test statistic, which contained pairwise terms (Eq.\,\ref{eq:mmd}), for large sample sizes. Furthermore, due to the poor shift detection performance of MMDD on retinal images in preliminary experiments, we replaced the shallow convolutional network in the original feature extractor $f_{\mbox{MMDD}}$ with a ResNet-50 model with a $128$-dimensional output. 

Classifier-based tests consistently outperformed the other hypothesis tests on all subgroup shift settings.
For detecting a subgroup shift towards lower-quality images, C2ST was distinctly better than MMDD and MUKS (Fig.~\ref{fig:eyepacs-main-results}\,\textbf{a}). On the other hand, subgroup shift in the prevalence of co-morbidities was easier to detect for all methods (Fig.~\ref{fig:eyepacs-main-results}\,\textbf{b}). While C2ST again outperformed the other methods, MUKS also performed notably well on this task. This could be explained by the properties of the MUKS test, which aimed to detect systematic differences in softmax predictions of a DR grading model. It is possible that these DR grade predictions could be influenced by co-morbidities, thus altering the distributions being measured by the MUKS method. Shifts in patient sex were notably more difficult to detect for all methods, with MMDD and MUKS failing completely (Fig.~\ref{fig:eyepacs-main-results}\,\textbf{c}). For ethnicity shifts, the shift detection rate varied between ethnicities (shaded areas in Fig.~\ref{fig:eyepacs-main-results}\,\textbf{d}), but overall, again MUKS yielded less reliable shift detection, and C2ST outperformed the others by a large margin. For all tests, consistent with previous experiments, the type I error remained approximately within the chosen significance level (Fig.~\ref{fig:eyepacs-main-results}\,\textbf{e}).

\subsubsection{Influence of image resolution on test power}
\label{sec:mmdd-ablation}

It is well-known that automated DR grading from fundus images requires high resolution fundus images \cite{huang2021identifying}. This motivated the use of $512 \times 512$ sized images wherever possible. To explain whether the large performance gap between MMDD and C2ST could be explained by the reduced image size for MMDD, we evaluated C2ST at lower resolution (C2ST-96) as well. Interestingly, reducing the image size from $512$ to $96$ only marginally degraded the subgroup shift detection performance for most shifts (Table\,\ref{tab:ablation-image-size}). The notable exception was a shift in patient sex, where test power was highly sensitive to image size. Overall, C2ST-96 consistently outperformed MMDD-96, indicating that image resolution was not a key reason for MMDD's inferior performance.

\begin{table}[b]
\centering
\caption{Influence of input image resolution on C2ST and MMDD test power for subgroup shifts in retinal images.}
\label{tab:ablation-image-size}
\begin{tabular}{llllllll}
\hline 
\multirow{2}{*}{Shift} & \multirow{2}{*}{} & \multicolumn{6}{c}{Test sample size} \\\cline{3-8}
 &  & 10 & 30 & 50 & 100 & 200 & 500 \\
\hline 
\multirow{3}{*}{Quality} & C2ST-512 & 0.16 & 0.73 & 0.86 & 1.00 & 1.00 & 1.00\\  
 & C2ST-96 & 0.19 & 0.54 & 0.78 & 0.99 & 1.00 & 1.00\\  
 & MMDD-96 & 0.09 & 0.05 & 0.14 & 0.22 & 0.48 & 0.94\\  
\hline 
\multirow{3}{*}{Co-morbid.} & C2ST-512 & 0.68 & 1.00 & 1.00 & 1.00 & 1.00 & 1.00\\  
 & C2ST-96 & 0.46 & 1.00 & 1.00 & 1.00 & 1.00 & 1.00\\  
 & MMDD-96 & 0.12 & 0.25 & 0.38 & 0.80 & 1.00 & 1.00\\  
\hline 
\multirow{3}{*}{Sex} & C2ST-512 & 0.07 & 0.28 & 0.58 & 0.78 & 0.98 & 1.00\\  
 & C2ST-96 & 0.07 & 0.10 & 0.13 & 0.08 & 0.27 & 0.66\\  
 & MMDD-96 & 0.05 & 0.07 & 0.04 & 0.06 & 0.12 & 0.18\\  
\hline
\multirow{3}{*}{Ethnicity} & C2ST-512 & 0.79 & 0.91 & 0.93 & 0.99 & 1.00 & 1.00\\  
 & C2ST-96 & 0.70 & 0.88 & 0.92 & 0.98 & 1.00 & 1.00\\  
 & MMDD-96 & 0.20 & 0.50 & 0.65 & 0.80 & 0.86 & 0.90\\  
\hline 
\end{tabular}
\end{table}

\subsubsection{Influence of architecture on classifier-based tests}
\label{sec:c2st-arch-ablation}

To further probe the behaviour of classifier-based tests, we studied their robustness to less powerful domain classifier architectures using quality subgroup shifts as an example setting. We replaced the original ResNet-50 backbone in the domain classifier network with a ResNet-18 and also with a shallow model consisting of four convolutional layers similar to the architecture used in MMDD. Test power was reduced in both cases (Table\,\ref{tab:ablation-c2st-arch}). As expected, degradation was more notable with a shallow domain classifier, while the ResNet-18 only led to minor reductions in test power. The ResNet-18 was slightly better at $m=10$ ($0.22 \pm 0.13$), which could be explained by the large standard error for such a small sample size. 

\begin{table}[b]
\centering
\caption{Influence of the domain classifier architecture on test power for a quality shift in retinal fundus images.}
\label{tab:ablation-c2st-arch}
\begin{tabular}{lllllll}
\hline 
  & \multicolumn{6}{c}{Test sample size} \\\cline{2-7}
  & 10 & 30 & 50 & 100 & 200 & 500 \\
\hline 
C2ST-ResNet-50 & 0.16 & 0.73 & 0.86 & 1.00 & 1.00 & 1.00\\  
C2ST-ResNet-18 & 0.22 & 0.58 & 0.82 & 1.00 & 1.00 & 1.00\\  
C2ST-Shallow & 0.09 & 0.08 & 0.22 & 0.49 & 0.68 & 1.00 \\  
\hline 
\end{tabular}
\end{table}


\subsubsection{Influence of training dataset size on test power}
\label{sec:train-set-size-ablation}

We also studied the influence of the size of the training fold on test power, again using quality subgroup shifts as an example. We retrained all tests with exceedingly smaller fractions of the original training data, and evaluated test power at a fixed sample size of $m=100$. All test approaches were relatively robust to reducing the training set size to $10\%$ (Table\,\ref{tab:ablation-train-set-size}). Classifier-based tests still achieved a test power of $0.72$ with just $1\%$ of the original training data, while MMDD and MUKS degraded to $0.14$ and $0.03$.

\begin{table}[b]
\centering
\caption{Influence of the size of the training dataset on test power for a quality shift in retinal fundus images. The test sample size was set to $m=100$.}
\label{tab:ablation-train-set-size}
\begin{tabular}{lllll}
\hline 
  & \multicolumn{4}{c}{Training set size} \\\cline{2-5}
  & $100\%$ & $50\%$ & $10\%$ & $1\%$ \\
\hline 
C2ST & 1.00 & 0.98 & 0.95 & 0.72\\  
MMDD & 0.22 & 0.24 & 0.24 & 0.14\\  
MUKS & 0.24 & 0.23 & 0.25 & 0.03\\  
\hline 
\end{tabular}
\end{table}

\subsubsection{Sensitivity to subgroup shift strength}

\begin{figure}
\centering{\includegraphics{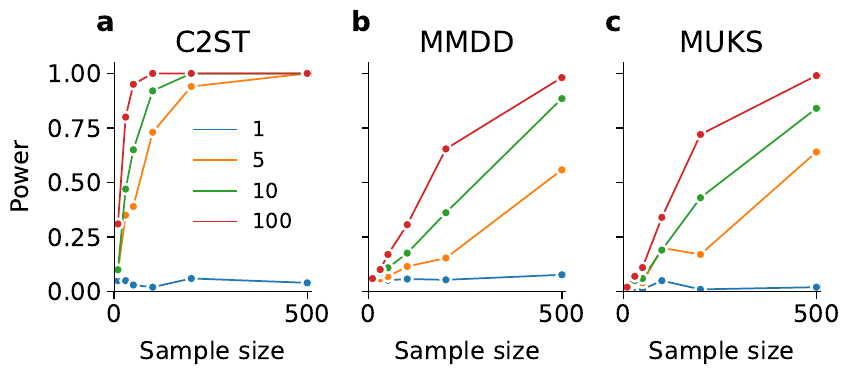}}
\caption{Influence of varying quality subgroup shift strength on test power of \textbf{a)} C2ST, \textbf{b)} MMDD, and \textbf{c)} MUKS tests.}
\label{fig:ablation-eyepacs-gradual}
\end{figure}

Finally, we examined the sensitivity of our subgroup shift detection tests to only subtle changes between distributions $\mathbb{P}$ and $\mathbb{Q}$. To control the subgroup shift strength, we artificially adjusted the prevalence of low quality images. We sampled with replacement from the whole dataset, and gradually over-represented these images with factors $w=\{1, 5, 10, 100\}$ in $\mathbb{Q}$ compared to $\mathbb{P}$. As expected, stronger shifts led to higher detection rate (Table\,\ref{fig:ablation-eyepacs-gradual}).Interestingly, classifier-based tests detected shifts reliably even with a small over-sampling factor $w=5$, while performance differences were more severe for MMDD and MUKS. No oversampling ($w=1$, blue curves) implied $\mathbb{P} = \mathbb{Q}$, and there test power expectedly remained in line with the chosen significance level.

\section{Discussion and Conclusion}

In this paper, we studied subgroup shift detection for a variety of distribution shifts and on different image datasets. We first showed that subgroup shifts can indeed affect the overall performance of a ML system by reporting subgroup performance disparities for various classification tasks. 
For example, we have shown that in retinal fundus images, the subgroup shifts we modelled led to an overall accuracy drop of up to $0.06$ (from $0.87$ to $0.81$).
We then adopted three state-of-the-art approaches towards detecting subgroup shifts in medical images. All approaches used the framework of statistical hypothesis testing for shift detection and relied on neural network based representations of the images for meaningful test statistics. Our experiments showed that in particular classifier-based tests (C2ST) consistently and considerably outperformed the other approaches. 

In our earlier work \cite{koch2022hidden}, we had omitted C2ST from our experiments as \cite{liuLearningDeepKernels} had found them to be inferior to MMDD for detecting distribution shifts on less complex data distributions. The results we reported here contradict these previous findings. We suspect that C2ST yielded superior power because the C2ST test relies on training a ``vanilla'' classification task (i.e. a binary domain classifier), and as a community, we have converged to a good understanding of suitable hyperparameters and architectural choices for solving such tasks. In contrast, training an MMDD test requires maximising the maximum mean discrepancy between two minibatches, and more exploration may be need to solve this task robustly.

Multiple univariate Kolmogorov-Smirnov tests \cite{rabanser2019FailingLoudlyEmpirical} yielded lower test power in most subgroup shift configurations. Since MUKS tests rely on a task classifier as a feature extractor, MUKS performance is by design tied to systematic changes in task predictions. The MUKS test therefore seemed to work reasonably well in scenarios such as co-morbidity shifts, where it is plausible that task predictions were affected. While not ideal in terms of general test power, MUKS tests have the advantage that they do not require training data from the deployment distribution. Depending on the availability of real-world data, this may make MUKS more applicable than C2ST and MMDD, which are both optimised on training data from both the clinical validation and deployment setting. However, as C2ST and MMDD tests are completely task-agnostic and do not require labelled data for training, the availability of real-world data should not pose a hurdle in many settings.

Overall, we have shown that effective tools exist for detecting clinically relevant subgroup distribution shifts, which (as demonstrated in \cite{koch2022hidden}) cannot be identified with OOD detection methods. While here, for the sake of demonstration, we relied on labelled attributes to simulate subgroup shift settings, we emphasise again that the proposed tools will be applied in a setting where we do not know such attributes. Rather, our approach requires only access to unlabelled samples from two distributions (i.e. validation and deployment data).

We therefore see the proposed subgroup detection techniques as useful and easily implemented components in the post-market surveillance strategy of medical device manufacturers. While detected distribution shifts would not automatically imply a performance drop of the ML system, such events could trigger an investigation into the root cause and potential consequences of the detected shift.



\appendices


\section*{Acknowledgments}

We thank Sarah M{\"u}ller for providing data preprocessing tools and Valentyn Boreiko and Jonas K{\"u}bler for helpful discussions.

\bibliographystyle{IEEEtran}
\bibliography{references}

\begin{thebibliography}{10}
\providecommand{\url}[1]{#1}
\csname url@samestyle\endcsname
\providecommand{\newblock}{\relax}
\providecommand{\bibinfo}[2]{#2}
\providecommand{\BIBentrySTDinterwordspacing}{\spaceskip=0pt\relax}
\providecommand{\BIBentryALTinterwordstretchfactor}{4}
\providecommand{\BIBentryALTinterwordspacing}{\spaceskip=\fontdimen2\font plus
\BIBentryALTinterwordstretchfactor\fontdimen3\font minus
  \fontdimen4\font\relax}
\providecommand{\BIBforeignlanguage}[2]{{%
\expandafter\ifx\csname l@#1\endcsname\relax
\typeout{** WARNING: IEEEtran.bst: No hyphenation pattern has been}%
\typeout{** loaded for the language `#1'. Using the pattern for}%
\typeout{** the default language instead.}%
\else
\language=\csname l@#1\endcsname
\fi
#2}}
\providecommand{\BIBdecl}{\relax}
\BIBdecl

\bibitem{liu2019ComparisonDeepLearning}
X.~Liu, L.~Faes, A.~U. Kale, S.~K. Wagner, D.~J. Fu, A.~Bruynseels,
  T.~Mahendiran, G.~Moraes, M.~Shamdas, C.~Kern, J.~R. Ledsam, M.~K. Schmid,
  K.~Balaskas, E.~J. Topol, L.~M. Bachmann, P.~A. Keane, and A.~K. Denniston,
  ``A comparison of deep learning performance against health-care professionals
  in detecting diseases from medical imaging: a systematic review and
  meta-analysis,'' \emph{The Lancet Digital Health}, 2019.

\bibitem{mdr2017}
{European Parliament, Council of the European Union}, ``Regulation ({EU})
  2017/745 of the {European Parliament} and of the {Council} of 5 april 2017 on
  medical devices, amending {Directive 2001/83/EC, Regulation (EC) No 178/2002
  and Regulation (EC) No 1223/2009} and repealing {Council Directives
  90/385/EEC and 93/42/EEC},'' \emph{Official Journal of the European Union},
  2017.

\bibitem{eu2021draftai}
{European Commission, Directorate-General for Communications Networks, Content
  and Technology }, ``{Proposal for a Regulation of the European Parliament and
  of the Council laying down harmonised rules on artificial intelligence
  (Artificial Intelligence Act) and amending certain union legislative acts},''
  2021.

\bibitem{FDA2021}
U.~Food and D.~A. (FDA), ``Artificial {I}ntelligence/{M}achine learning
  {(AI/ML)}-{B}ased {S}oftware as a {M}edical {D}evice {(SaMD)} {A}ction
  {P}lan,'' 2021, \url{https://www.fda.gov/media/145022/download}.

\bibitem{LIU2022e384}
X.~Liu, B.~Glocker, M.~M. McCradden, M.~Ghassemi, A.~K. Denniston, and
  L.~Oakden-Rayner, ``The medical algorithmic audit,'' \emph{The Lancet Digital
  Health}, vol.~4, no.~5, pp. e384--e397, 2022.

\bibitem{finlayson2021ClinicianDatasetShift}
S.~G. Finlayson, A.~Subbaswamy, K.~Singh, J.~Bowers, A.~Kupke, J.~Zittrain,
  I.~S. Kohane, and S.~Saria, ``The {Clinician} and {Dataset} {Shift} in
  {Artificial} {Intelligence},'' \emph{New England Journal of Medicine}, vol.
  385, no.~3, pp. 283--286, Jul. 2021.

\bibitem{oakden2020hiddenStratification}
L.~Oakden-Rayner, J.~Dunnmon, G.~Carneiro, and C.~Re, ``Hidden stratification
  causes clinically meaningful failures in machine learning for medical
  imaging,'' in \emph{Proceedings of the ACM Conference on Health, Inference,
  and Learning}, 2020.

\bibitem{eyuboglu2022domino}
S.~Eyuboglu, M.~Varma, K.~K. Saab, J.-B. Delbrouck, C.~Lee-Messer, J.~Dunnmon,
  J.~Zou, and C.~Re, ``Domino: Discovering systematic errors with cross-modal
  embeddings,'' in \emph{International Conference on Learning Representations},
  2022.

\bibitem{koch2022hidden}
L.~M. Koch, C.~M. Sch{\"u}rch, A.~Gretton, and P.~Berens, ``Hidden in plain
  sight: Subgroup shifts escape {OOD} detection,'' in \emph{Proc. Medical
  Imaging with Deep Learning (MIDL)}, 2022.

\bibitem{liuLearningDeepKernels}
F.~Liu, W.~Xu, J.~Lu, G.~Zhang, A.~Gretton, and D.~J. Sutherland, ``Learning
  {Deep} {Kernels} for {Non}-{Parametric} {Two}-{Sample} {Tests},'' in
  \emph{Proc. International Conference on Machine Learning (ICML)}, 2020.

\bibitem{rabanser2019FailingLoudlyEmpirical}
S.~Rabanser, S.~Günnemann, and Z.~C. Lipton, ``Failing {Loudly}: {An}
  {Empirical} {Study} of {Methods} for {Detecting} {Dataset} {Shift},'' in
  \emph{Proc. Advances in Neural Information Processing Systems (NeurIPS)},
  2019.

\bibitem{cheng2019classification}
X.~Cheng and A.~Cloninger, ``Classification logit two-sample testing by neural
  networks,'' \emph{arXiv preprint arXiv:1909.11298}, 2019.

\bibitem{shimodaira2000improving}
H.~Shimodaira, ``Improving predictive inference under covariate shift by
  weighting the log-likelihood function,'' \emph{Journal of statistical
  planning and inference}, vol.~90, no.~2, pp. 227--244, 2000.

\bibitem{Sagawa2020Distributionally}
S.~Sagawa, P.~W. Koh, T.~B. Hashimoto, and P.~Liang, ``Distributionally robust
  neural networks,'' in \emph{International Conference on Learning
  Representations}, 2020.

\bibitem{cui2019classbalanced}
Y.~Cui, M.~Jia, T.-Y. Lin, Y.~Song, and S.~Belongie, ``Class-balanced loss
  based on effective number of samples,'' in \emph{Proceedings of the IEEE/CVF
  Conference on Computer Vision and Pattern Recognition (CVPR)}, 2019.

\bibitem{wilds2021}
P.~W. Koh, S.~Sagawa, H.~Marklund, S.~M. Xie, M.~Zhang, A.~Balsubramani, W.~Hu,
  M.~Yasunaga, R.~L. Phillips, I.~Gao, T.~Lee, E.~David, I.~Stavness, W.~Guo,
  B.~A. Earnshaw, I.~S. Haque, S.~Beery, J.~Leskovec, A.~Kundaje, E.~Pierson,
  S.~Levine, C.~Finn, and P.~Liang, ``{WILDS}: A benchmark of in-the-wild
  distribution shifts,'' in \emph{Proc. International Conference on Machine
  Learning (ICML)}, 2021.

\bibitem{jain2023distilling}
S.~Jain, H.~Lawrence, A.~Moitra, and A.~Madry, ``Distilling model failures as
  directions in latent space.''

\bibitem{kubler2022automl}
J.~M. K{\"u}bler, V.~Stimper, S.~Buchholz, K.~Muandet, and B.~Sch{\"o}lkopf,
  ``Automl two-sample test,'' \emph{arXiv preprint arXiv:2206.08843}, 2022.

\bibitem{lopez2016revisiting}
D.~Lopez-Paz and M.~Oquab, ``Revisiting classifier two-sample tests,'' in
  \emph{International Conference on Learning Representations (ICLR)}, 2017.

\bibitem{yang2022openood}
J.~Yang, P.~Wang, D.~Zou, Z.~Zhou, K.~Ding, W.~PENG, H.~Wang, G.~Chen, B.~Li,
  Y.~Sun, X.~Du, K.~Zhou, W.~Zhang, D.~Hendrycks, Y.~Li, and Z.~Liu,
  ``Open{OOD}: Benchmarking generalized out-of-distribution detection,'' in
  \emph{Proc. Neural Information Processing Systems: Datasets and Benchmarks
  Track}, 2022.

\bibitem{casella2002statistical}
G.~Casella and R.~Berger, \emph{Statistical Inference}.\hskip 1em plus 0.5em
  minus 0.4em\relax Duxbury, 2002.

\bibitem{gretton2012KernelTwoSampleTest}
A.~Gretton, K.~Borgwardt, M.~J. Rasch, B.~Schoelkopf, and A.~Smola, ``A
  {Kernel} {Two}-{Sample} {Test},'' \emph{The Journal of Machine Learning
  Research}, 2012.

\bibitem{sutherland2021GenerativeModelsModel}
D.~J. Sutherland, H.-Y. Tung, H.~Strathmann, S.~De, A.~Ramdas, A.~Smola, and
  A.~Gretton, ``Generative {Models} and {Model} {Criticism} via {Optimized}
  {Maximum} {Mean} {Discrepancy},'' in \emph{Proc. International Conference on
  Learning Representations (ICLR)}, 2017.

\bibitem{lecun1998mnist}
Y.~Lecun, L.~Bottou, Y.~Bengio, and P.~Haffner, ``Gradient-based learning
  applied to document recognition,'' \emph{Proceedings of the IEEE}, 1998.

\bibitem{kingma2015adam}
D.~P. Kingma and J.~Ba, ``Adam: A method for stochastic optimization,'' in
  \emph{International Conference on Learning Representations (ICLR)}, 2015.

\bibitem{bandi2019camelyon}
P.~Bándi, O.~Geessink, Q.~Manson, M.~Van~Dijk, M.~Balkenhol, M.~Hermsen,
  B.~Ehteshami~Bejnordi, B.~Lee, K.~Paeng, A.~Zhong, Q.~Li, F.~G. Zanjani,
  S.~Zinger, K.~Fukuta, D.~Komura, V.~Ovtcharov, S.~Cheng, S.~Zeng,
  J.~Thagaard, A.~B. Dahl, H.~Lin, H.~Chen, L.~Jacobsson, M.~Hedlund,
  M.~Çetin, E.~Halıcı, H.~Jackson, R.~Chen, F.~Both, J.~Franke,
  H.~Küsters-Vandevelde, W.~Vreuls, P.~Bult, B.~van Ginneken, J.~van~der Laak,
  and G.~Litjens, ``From detection of individual metastases to classification
  of lymph node status at the patient level: The camelyon17 challenge,''
  \emph{IEEE Transactions on Medical Imaging}, 2019.

\bibitem{international2017ico}
ICO, ``International council of ophthalmology (ico) guidelines for diabetic eye
  care,'' 2017.

\bibitem{mueller2022a}
S.~M{\"u}ller, L.~M. Koch, H.~Lensch, and P.~Berens, ``A generative model
  reveals the influence of patient attributes on fundus images,'' in
  \emph{Proc. Medical Imaging with Deep Learning -- short}, 2022.

\bibitem{huang2021identifying}
Y.~Huang, L.~Lin, P.~Cheng, J.~Lyu, and X.~Tang, ``Identifying the key
  components in resnet-50 for diabetic retinopathy grading from fundus images:
  a systematic investigation,'' \emph{arXiv:2110.14160}, 2021.

\end{thebibliography}

\end{document}